\documentclass{tlp}

\usepackage{amsmath}
\usepackage{graphicx}
\usepackage{multirow}
\usepackage{hyperref}

\usepackage{graphicx}        
\usepackage{multicol}        
\usepackage[bottom]{footmisc}

\usepackage{color}
\usepackage{array}
\usepackage{tabularx}
\usepackage{url}
\usepackage{amsmath}
\usepackage{xspace}
\usepackage{paralist}
\usepackage{relsize} 
\usepackage{subcaption}
\usepackage{subfloat}
\usepackage{algorithm}
\usepackage{algpseudocode}
\usepackage{amssymb}
\usepackage{pdfsync}

\usepackage{thm-restate}
\declaretheorem[numberwithin=section]{example}

\newenvironment{monospace}{\fontfamily{qcr}\selectfont}{\par}

\def\UF{\ensuremath{AF}}
\newcommand{\SF}{\ensuremath{PF}}
\newcommand{\OF}{\ensuremath{O\!F}}
\newcommand{\NF}{\ensuremath{N\!F}}

\newcommand{\NR}{\ensuremath{N\!R}}

\newcommand{\DG}{\ensuremath{DG}}

\newcommand{\veel}{ \ |\ }

\newcommand{\nop}[1]{}

\def\embeds{\vdash}

\def\tailors{\Vdash}

\def\deltainst{\ensuremath{\Delta{{\textsc{{Inst}}}}}\xspace}

\def\incrinst{\ensuremath{\textsc{{IncrInst}}}\xspace}
\def\desimpl{\textsc{Desimpl}\xspace}

\newcommand{\sysfont}{\textit}

\newcommand{\dlv}{\sysfont{DLV}\xspace}
\newcommand{\dlvdue}{\sysfont{DLV\smaller[0.8]{2}}\xspace}

\newcommand{\system}{\sysfont{Incremental-DLV\smaller[0.8]{2}}\xspace}
\newcommand{\itwodlv}{{\mbox{\small ${I}^2$-\dlv}}\xspace}

\def\pvsystem{\textit{Photo-voltaic System}\xspace}
\def\pacman{\textit{Pac-Man}\xspace}
\def\mmedia{\textit{Content Caching}\xspace}

\newcommand\derives{\ensuremath{\scalebox{0.6}[1.0]{\(:\!\!- \)}}}

\newcommand{\naf}{\ensuremath{\mathtt{not}}\xspace}

\newcommand{\cit}[1]{~\cite{#1}}
\newcommand{\re}[1]{~\ref{#1}}

\newcommand\quo[1]{``#1''}

\newcommand\hcancel[2][black]{\setbox0=\hbox{$#2$}%
\rlap{\raisebox{.30\ht0}{\textcolor{#1}{\rule{\wd0}{0.5pt}}}}#2}

\newif\ifrevisionmode
\revisionmodefalse

\ifrevisionmode
    
    \newcommand{\remove}[1] { {\color{red}\sout{#1}} }

    \newcommand{\TODO}[1]{{\color{red}{TODO: #1}}}
    \newcommand{\oldstuff}[1]{{\color{gray}FROM OLD PAPER: #1}}

\else

    \newcommand{\remove}[1]{}

    \newcommand{\TODO}[1]{}
    \newcommand{\oldstuff}[1]{{}}

\fi

\begin{document}

\lefttitle{Calimeri et al.}

\jnlPage{X}{Y}
\jnlDoiYr{2024}
\doival{10.1017/xxxxx}

\title[ASP-based Multi-shot Reasoning]{ASP-based Multi-shot Reasoning via DLV2 with Incremental Grounding\thanks{This work has been partially supported by the Italian MIUR Ministry and the Presidency of the Council of Ministers under the project \quo{Declarative Reasoning over Streams} under the \quo{PRIN} 2017 call (CUP $H24I17000080001$, project 2017M9C25L\_001); by the Italian Ministry of Economic Development (MISE) under the PON  project \quo{MAP4ID - Multipurpose Analytics Platform 4 Industrial Data}, N. {F/190138/01-03/X44}; and by European Union under the National Recovery and Resilience Plan (NRRP) funded by the italian MUR ministry with the project Future Artificial Intelligence Research (FAIR, PE0000013), Spoke 9. 
Francesco Calimeri is member of the Grup\-po Na\-zio\-na\-le Calcolo Scientifico-Istituto Na\-zio\-nale di Alta Matematica (GNCS-INdAM). 
Competing interests: Francesco Calimeri is CEO and Sole director of the Italian limited liability company (LLC) {\em DLVSystem Srl} and owns shares of the company; Simona Perri owns shares of {\em DLVSystem Srl}; other authors declare none. 
{\em DLVSystem Srl} is a spin-off company of University of Calabria.
A shorter version of this paper has been presented at PPDP 2022: 24th International Symposium on Principles and Practice of Declarative Programming.}}

\def\unical{University of Calabria}
\def\dlvsystem{DLVSystem Srl}

\begin{authgrp}
\author{\sn{Francesco} \gn{Calimeri}}
\affiliation{\unical}
\affiliation{\dlvsystem}
\author{\sn{Giovambattista} \gn{Ianni}}
\affiliation{\unical}
\author{\sn{Francesco} \gn{Pacenza}}
\affiliation{\unical}
\author{\sn{Simona} \gn{Perri}}
\affiliation{\unical}
\author{\sn{Jessica} \gn{Zangari}}
\affiliation{\unical}
\end{authgrp}

\maketitle

\begin{abstract}

DLV2 is an AI tool for Knowledge Representation and Reasoning which supports Answer Set Programming (ASP) -- a logic-based declarative formalism, successfully used in both academic and industrial applications. 
Given a logic program modelling a computational problem, an execution of DLV2 produces the so-called answer sets that correspond one-to-one to the solutions to the problem at hand. 
The computational process of DLV2 relies on the typical Ground \& Solve approach where the grounding step transforms the input program into a new, equivalent ground program, and the subsequent solving step applies propositional algorithms to search for the answer sets.
Recently, emerging applications in contexts such as stream reasoning and event processing created a demand for multi-shot reasoning: here, the system is expected to be reactive while repeatedly executed over rapidly changing data.
In this work, we present a new incremental reasoner obtained from the evolution of DLV2 towards iterated reasoning.
Rather than restarting the computation from scratch, the system remains alive across repeated shots, and it incrementally handles the internal grounding process. At each shot, the system reuses previous computations for building and maintaining a large, more general ground program, from which a smaller yet equivalent portion is determined and used for computing answer sets. Notably, the incremental process is performed 
in a completely transparent fashion for the user.
We describe the system, its usage, its applicability and performance in some practically relevant domains.

Under consideration in Theory and Practice of Logic Programming (TPLP).
\end{abstract}



\maketitle


\section{Introduction}
Answer Set Programming (ASP) is a declarative problem-solving formalism that emerged in the area of logic programming and nonmonotonic reasoning~(\cite{DBLP:journals/cacm/BrewkaET11,gelf-lifs-1991,DBLP:conf/rweb/EiterIK09-long}).
%
Thanks to its solid theoretical foundations and the availability of efficient implementations~(see \cite{DBLP:conf/ijcai/GebserLMPRS18} for a survey), ASP is recognized as a powerful tool for Knowledge Representation and Reasoning (KRR) and became widely used in AI. 

Rules represent the basic linguistic construct in ASP. 
A rule has form $Head\leftarrow Body$, where the $Body$ is a logic conjunction in which negation may appear, and $Head$ can be either an atomic formula or a logic disjunction; 
a rule is interpreted according to common sense principles: roughly, its intuitive semantics corresponds to an implication.
Rules featuring an atomic formula in the head and an empty body are used to represent information known to be certainly true and are indeed called facts. 
In ASP, a computational problem is typically solved by modelling it via a logic program consisting of a collection of rules along with a set of facts representing the instance at hand, and then by making use of an ASP system that determines existing solutions by computing the intended models, called \textit{answer sets}. The latter are computed according to the so-called \textit{answer set semantics}. 
Answer sets correspond one-to-one to the solutions of the given instance of the modeled problem; if a program has no answer sets, the corresponding problem instance has no solutions.

The majority of currently available ASP systems relies on the traditional \quo{Ground \& Solve} workflow which is based on two consecutive steps.
First, a grounding  step (also said instantiation step) transforms the input program into a semantically equivalent ``ground'' program, i.e., a propositional program without first-order variables. 
Then, in a subsequent solving step, algorithms are applied on this ground program to compute the corresponding answer sets.
There are other systems which, instead, are based on approaches that interleave grounding and solving or rely on intermediate translations like the ones presented in ~\cite{DBLP:conf/aaai/BomansonJW19},\cite{DBLP:journals/fuin/PaluDPR09}, and \cite{DBLP:journals/tplp/LefevreBSG17}.


In the latest years, emerging application contexts, such as real-time motion tracking (\cite{DBLP:conf/aaai/SuchanBWS18}), content distribution (\cite{DBLP:conf/icc/BeckBDEHS17}),  robotics (\cite{DBLP:journals/arobots/SaribaturPE19}), artificial players in videogames (\cite{DBLP:conf/ruleml/CalimeriGIPPZ18}), and sensor network configuration (\cite{DBLP:journals/tplp/DodaroEOS20}), 
have been posing new challenges for ASP systems. 
Most of the above applications require to show high reactivity while performing the repeated execution of reasoning tasks over rapidly changing input data. Each repeated execution is commonly called ``shot'', hence the terminology \quo{multi-shot} reasoning.
%
In the context of multi-shot reasoning, the na\"ive approach of starting ASP systems at hand from scratch at each execution significantly impacts on performance, and is impracticable when shots are needed at a very high pace and/or over a high volume of input data. 


Lately, many efforts have been spent by the scientific community to define proper incremental evaluation techniques that save and reuse knowledge built across shots, thus 
making ASP systems and general rule-based systems evolve towards more efficient multi-shot solutions, such as the works of~\cite{DBLP:journals/ai/MotikNPH19}, \cite{DBLP:journals/tplp/GebserKKS19}, \cite{DBLP:journals/datasci/DellAglioVHB17}, \cite{DBLP:conf/rr/MileoAPH13}, \cite{DBLP:conf/fis/ValleCBBC08}, \cite{DBLP:journals/tplp/GebserKKS19}, \cite{DBLP:journals/tplp/CalimeriIPPZ19}, \cite{DBLP:journals/tplp/IanniPZ20}, and \cite{DBLP:journals/tplp/BeckEB17}.

\smallskip

In this work, we present \system, a new incremental ASP reasoner that represents the evolution of \dlvdue of~\cite{DBLP:conf/lpnmr/AlvianoCDFLPRVZ17} towards multi-shot reasoning.
\dlvdue is a novel version of one of the first and more widespread ASP system, namely \dlv (\cite{DBLP:journals/tocl/LeonePFEGPS06}); the new system has been re-implemented from scratch and encompasses the outcome of the latest research effort on both grounding and solving areas.
Just as \dlv and \dlvdue, \system entirely embraces the declarative nature of ASP; furthermore, it contributes to research in multi-shot solving with the introduction and management of a form of incremental grounding which is fully transparent to users of the system. This is achieved via the overgrounding techniques presented in~\cite{DBLP:journals/tplp/CalimeriIPPZ19}, and \cite{DBLP:journals/tplp/IanniPZ20}.
The overgrounding approach makes, at each shot, the instantiation effort directly proportional to the number of unseen facts, up to the point that the grounding computational effort might be close to none when all input facts have been already seen in previous shots. 
Notably, overgrounded programs are increasingly larger across shots: as this could negatively impact on the solving step, \system properly selects only a smaller yet equivalent portion of the current overgrounded program to be considered during solving. 

\smallskip

In the remainder of the manuscript, we first provide an overview of the incremental grounding techniques, which the system relies on, in Section~\ref{sec:overgrounding}; then, we illustrate the system architecture and its computational workflow in Section~\ref{sec:system}; furthermore, we describe its usage and its applicability in Section~\ref{sec:usage}, while we assess the performance of the system in some practically relevant domains in Section~\ref{sec:exp+disc}.
Eventually, we discuss related work in Section~\ref{sec:relwork} and we conclude commenting about future work in Section~\ref{sec:conclusions}.

\section{Overview of Overgrounding Techniques}\label{sec:overgrounding}

In the following, we give an overview of the approach adopted by the system to efficiently manage the grounding task in multi-shot contexts.
We assume that the reader is familiar with the basic logic programming terminology, including the notions of predicate, atom, literal, rule, head, body, and refer to the literature for a detailed and systematic description of the ASP language and semantics (\cite{DBLP:journals/tplp/CalimeriFGIKKLM20}).



As mentioned, ASP solvers generally deal with a non-ground ASP program $P$, made of a set of universally quantified rules, and a set of input facts $F$. A traditional ASP system performs two separate steps to determine the corresponding models, i.e. the answer sets of $P$ and $F$, denoted $AS(P\cup F)$.
The first step is called {\em instantiation} (or {\em grounding}) and consists in the generation of a logic program $gr(P,F)$, obtained by properly replacing first-order variables with constants. 
Secondly, the solving step is responsible for computing the answer sets $AS(gr(P,F))$.
Grounding modules are typically geared towards building $gr(P,F)$ as a smaller and optimized version of the theoretical instantiation $grnd(P,F)$, which is classically defined via the Herbrand base.

When building $gr(P,F)$, it is implicitly assumed a ``one-shot'' context: the instantiation procedure is performed once and for all.
Hence, state-of-the-art grounders adopt ad-hoc strategies in order to heavily reduce the size of $gr(P,F)$.
In other words, $gr(P,F)$ is shaped on the basis of the problem instance at hand, still keeping its semantics. Basic equivalence is guaranteed as $gr$ is built in a way such that $AS(P \cup F) = AS(grnd(P,F)) = AS(gr(P,F))$.

Based on the information about the structure of the program and the given input facts, the generation of a significant number of useless ground rules can be avoided: for instance, rules having a definitely false literal in the body can be eliminated. Moreover, while producing a ground rule, on-the-fly simplification strategies can be applied; e.g., certainly true literals can be removed from rule bodies.
For an overview of grounding optimizations the reader can refer to~\cite{DBLP:conf/lpnmr/GebserKKS11}, \cite{DBLP:journals/ia/CalimeriFPZ17}, and \cite{cali-perr-zang-TPLP-optimizing}.

However, this optimization process makes $gr(P,F)$ \quo{tailored} for the $P \cup F$ input only. Assuming that $P$ is kept fixed, it is not guaranteed that, for a future different input $F'$, we will have $gr(P,F) = AS(P \cup F')$. 
Nonetheless, it might be desirable to maintain $gr(P,F)$ and incrementally modify it, with as little effort as possible, in order to regain equivalence for a subsequent shot with input set of facts $F'$. 

In this scenario it is crucial to limit as much as possible the regeneration of parts of the ground programs which were already evaluated at the previous step; at the same time, given that the set of input facts is possibly different from any other shot, shaping the produced ground program cannot be strongly optimized and tailored to $F'$ as in the ``one shot'' scenario. 
As a consequence, it is desirable that the instantiation process takes into account facts from both the current and the previous shots.
In this respect, \cite{DBLP:journals/tplp/CalimeriIPPZ19}, and \cite{DBLP:journals/tplp/IanniPZ20} proposed overgrounding techniques
to efficiently perform incremental instantiations.

The basic idea of the technique, originally introduced by~\cite{DBLP:journals/tplp/CalimeriIPPZ19}, is to maintain an  {\em overgrounded program} $G$. $G$ is monotonically enlarged across shots, in order to be semantics-preserving with respect to new input facts. 
Interestingly, the overgrounded version of $G$ resulting at a given iteration $i$ is semantics-preserving for all the set of input facts at a previous iteration $i'$ ($1 \leq i' \leq i$), still producing the correct answer sets. More formally, for each $i', (1\leq i' \leq i)$, we have $AS(G \cup F_{i'}) = AS(P \cup F_{i'})$
After some iterations, $G$ {\em converges} to a propositional theory that is general enough to be reused together with large families of possible future inputs, without requiring further updates.
In order to achieve the above property, $G$ is adjusted from one shot to another by adding new ground rules and avoiding specific input-dependent simplifications.
This virtually eliminates the need for grounding activities in later iterations, at the price of potentially increasing the burden of the solver (sub)systems, that are supposed to deal with larger ground programs.

Overgrounding with tailoring, proposed by~\cite{DBLP:journals/tplp/IanniPZ20}, has been introduced with the aim of overcoming such limitations by keeping the principle that $G$ grows monotonically from one shot to another yet adopting fine-tuned techniques that allow to reduce the number of additions to $G$ at each step. 
More in detail, in the overgrounding with tailoring approach, new rules added to $G$ are subject to simplifications, which cause the length of individual rules and the overall size of the overgrounded program to be reduced, but desimplifications are applied whenever necessary in order to maintain compatibility with input facts.

In the following, we illustrate how the two techniques behave across subsequent shots with the help of a proper example. 
\begin{example}\label{exmp:overgrounding}
Let us consider the program $P_{ex}$:
\begin{center}
\begin{tabular}{l}
$a : r(X,Y) \ \derives \ e(X,Y),\ \naf\ q(X).$ \\
$b : r(X,Z) \veel s(X,Z) \ \derives \ e(X,Y),\ r(Y,Z).$
\end{tabular}
\end{center}

Let us assume at shot $1$ to have the input facts $F_1 = \{ e(3,1),\ $ $e(1,2),\ $ $q(3)  \}$.
In the standard overgrounding approach we start from $F_1$ and generate, in a bottom-up way, new rules by iterating through positive body-head dependencies, obtaining the ground program $G_1$:

\begin{center}
\begin{tabular}{l}
$a_1 : r(1,2) \ \derives \ e(1,2),\ \naf\ q(1).$ \\
$b_1 : r(3,2)  \veel s(3,2) \ \derives \ e(3,1),\ r(1,2).$ \\
$a_2 : r(3,1) \ \derives \ e(3,1),\ \naf\ q(3).$
\end{tabular}
\end{center}

In the overgrounding with tailoring, rules that have no chance of firing along with definitely true atoms are simplified, thus obtaining a simplified program $G'_1$:
\begin{center}
\begin{tabular}{l}
$a'_1 : r(1,2) \ \derives \ \hcancel{e(1,2)},\ \naf\ q(1).$ \\
$b'_1 : r(3,2)  \veel s(3,2) \ \derives \ \hcancel{e(3,1)},\ r(1,2).$ \\
$a'_2 : \hcancel{r(3,1) \ \derives \ e(3,1),\ \naf\ q(3).}$
\end{tabular}
\end{center}

$G'_1$ can be seen as less general and \quo{re-usable} than $G_1$: $a'_1$ is simplified on the assumption that $e(1,2)$ will be always true, and $a'_2$ is deleted on the assumption that $q(3)$ is always true.

One might want to adapt $G'_1$ to be compatible with different sets of input facts, but this requires the additional effort of retracting no longer valid simplifications. 
In turn, enabling simplifications could improve solving performance since a smaller overgrounded program is built.

Let us now assume that the shot $2$ requires $P_{ex}$ to be evaluated over a different set of input facts $F_2 = \{ e(3,1),\ e(1,4),$\ $q(1) \}$. 
Note that, with respect to $F_1$, $F_2$ features the additions $F^+ = \{ e(1,4), q(1) \}$ and the deletions $F^- = \{ e(1,2), q(3)\}$.
In the standard overgrounding approach, since no simplification is done, $G_1$ can be easily adapted to the new input $F_2$ by incrementally augmenting it according to $F^+$; this turns into adding the following rules $\Delta G_1 = \{b_2, a_3\}$, thus obtaining $G_2$:
\begin{center}
\begin{tabular}{l}
$a_1 : r(1,2) \ \derives \ e(1,2),\ \naf\ q(1).$ \\
$b_1 : r(3,2)  \veel s(3,2) \ \derives \ e(3,1),\ r(1,2).$ \\
$a_2 : r(3,1) \ \derives \ e(3,1),\ \naf\ q(3).$\\
${\bf b_2 : r(3,4)  \veel s(3,4) \ \derives \ e(3,1),\ r(1,4).}$ \\
${\bf a_3 : r(1,4) \ \derives \ e(1,4),\ \naf\ q(1).}$
\end{tabular}
\end{center}

$G_2$ is equivalent to $P$, when evaluated over $F_1$ or $F_2$. Furthermore, $G_2$ enjoys the property of being compatible as it is, with every possible subset of $F_1 \cup F_2$.
%
In the case of overgrounding with tailoring, $G'_1$ needs to be re-adapted by undoing no longer valid simplifications.
In particular, rule $a_2$,  previously deleted since $q(3) \in F_1$, is now restored, given that $q(3) \not\in F_2$; moreover, rule $a'_1$ is reverted to its un-simplified version $a_1$, since $e(1,2) \not\in F_2$. 
$b'_1$ is left unchanged, as reasons that led to simplify $b_1$ into $b'_1$ are still valid (i.e., $e(3,1)$, featured in $F_1$, still appears in $F_2$).

The so-called {\em desimplification} step applied to $G'_1$ thus produces the following ground program:

\begin{center}
\begin{tabular}{l}
$a_1 : r(1,2) \ \derives \ e(1,2),\ \naf\ q(1).$ \\
$b'_1 : r(3,2)  \veel s(3,2) \ \derives \ \hcancel{e(3,1)},\ r(1,2).$ \\
$a_2 : r(3,1) \ \derives \ e(3,1),\ \naf\ q(3).$\\
\end{tabular}
\end{center}

A further incremental step then generates new ground rules $b_2$ and $a_3$, based on the presence of new facts $F^+$. Only newly generated rules are subject to simplifications according to $F_2$. 
In particular, $e(3,1)$ is simplified from the body of $b_2$, obtaining $b'_2$; $a_3$ is eliminated (i.e., an empty version $a'_3$ is generated) since $not\ q(1)$ is false; 
The resulting program $G'_2$ is as follows:
\begin{center}
\begin{tabular}{l}
$a_1 : r(1,2) \ \derives \ e(1,2),\ \naf\ q(1).$ \\
$b'_1 : r(3,2)  \veel s(3,2) \ \derives \ \hcancel{e(3,1)},\ r(1,2).$ \\
$a'_2 : r(3,1) \ \derives \ e(3,1),\ \naf\ q(3).$\\
${\bf b'_2 : r(3,4)  \veel s(3,4) \ \derives \ \hcancel{e(3,1)},\ r(1,4).}$ \\
${\bf a'_3 : \hcancel{r(1,4) \ \derives \ e(1,4),\ \naf\ q(1).}}$
\end{tabular}
\end{center}
It is worth noting that $G'_2$ maintains the same semantics of $P_{ex}$, when either $F_1$ or $F_2$ are given as input facts, but the semantics is not preserved with all possible subsets of $F_1 \cup F_2$.

If a third shot is requested over the input facts $F_3 = \{ e(1,4),$\ $e(3,1),$\ $e(1,2)\}$, we observe that $G_2$ does not need any further incremental update, as all facts in $F_3$ already appeared at previous steps; hence, $G_3 = G_2$. 

%
%
Concerning $G'_2$, the desimplification step reinstates $a'_3$ while no additional rules are generated in the incremental step. 
This leads to the ground program $G'_3$:
\begin{center}
\begin{tabular}{l}
$a'_1 : r(1,2) \ \derives \ e(1,2),\ \naf\ q(1).$ \\
$b'_1 : r(3,2)  \veel s(3,2) \ \derives \ \hcancel{e(3,1)},\ r(1,2).$ \\
$a'_2 : r(3,1) \ \derives \ e(3,1),\ \naf\ q(3).$\\
$b'_2 : r(3,4)  \veel s(3,4) \ \derives \ \hcancel{e(3,1)},\ r(1,4).$ \\
$a_3 : r(1,4) \ \derives \ e(1,4),\ \naf\ q(1).$
\end{tabular}
\end{center}

\end{example}

\section{The \system System}\label{sec:system}

In this Section we present the \system\ system, an incremental ASP reasoner stemming as a natural evolution of \dlvdue of~\cite{DBLP:conf/lpnmr/AlvianoCDFLPRVZ17} towards multi-shot incremental reasoning.
We first provide the reader with a general overview of the computational workflow, 
then discuss some insights about the main computational stages.

\subsection{Computational Workflow}\label{subsec:workflow}

\system is built upon a proper integration of the overgrounding-based incremental grounder \itwodlv, presented by \cite{DBLP:journals/tplp/IanniPZ20}, into \dlvdue.  
Coherently with its roots, \system fully complies with the declarative nature of ASP; 
among all the requirements, an important feature is that all means for enabling efficient multi-shot incremental reasoning are mostly transparent to the user. 
\system adapts the traditional ground \& solve pipeline that we briefly recalled in Section~\ref{sec:overgrounding} to the new incremental context. The grounding step of \system is based on overgrounding with tailoring; furthermore, in order to reduce the impact of a ground program that grows across steps, the solving phase selectively processes only a smaller, equivalent subset of the current overgrounded program. 

\begin{figure*}[t]
  \centering
    \includegraphics[width=.95\textwidth]{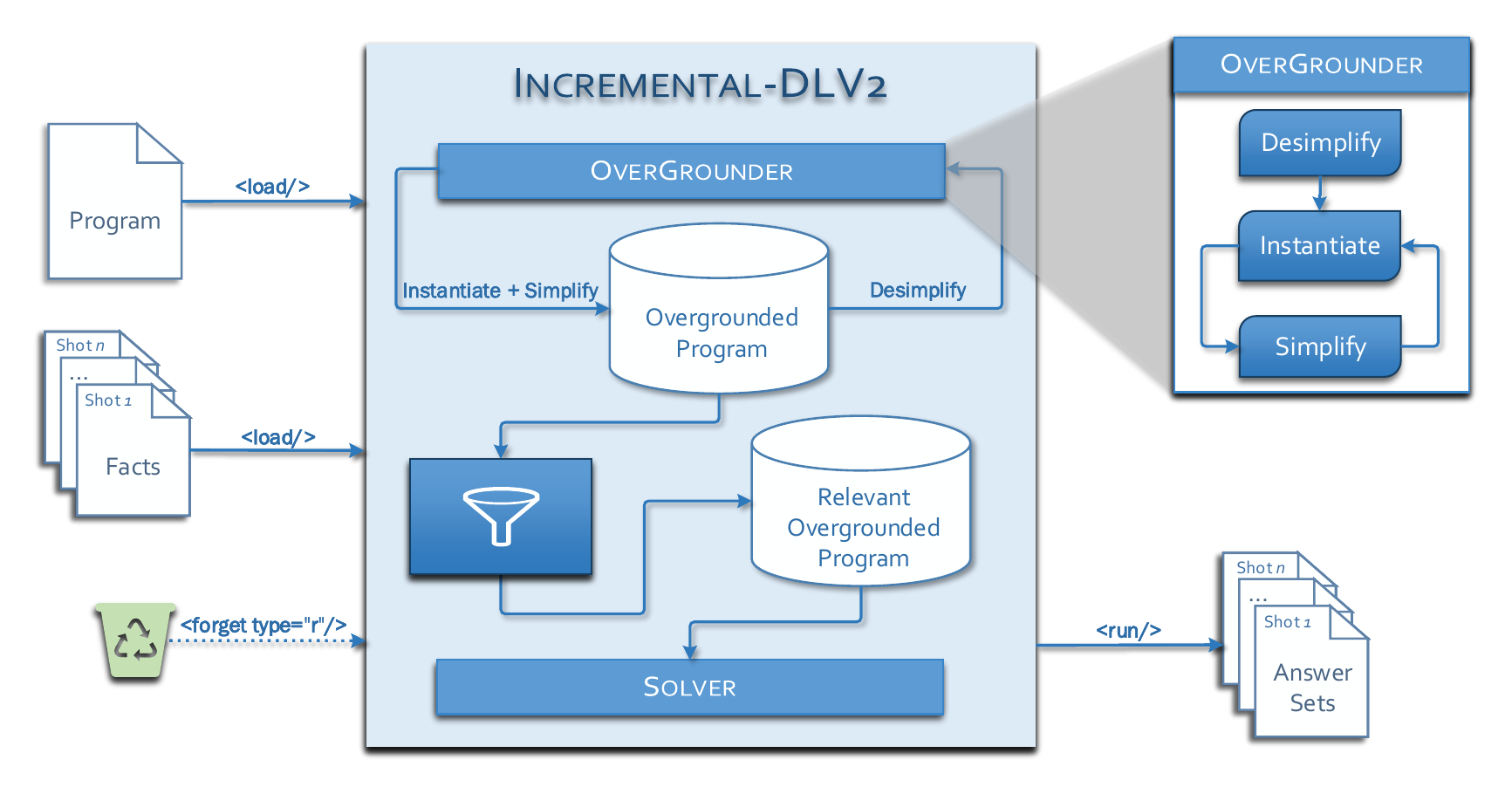}
  \caption{\system architecture.}
  \label{fig:architecture}
\end{figure*}

Figure~\ref{fig:architecture} provides a high-level picture of the internal workflow of the system. 
When \system is started, it keeps itself alive in a listening state waiting for commands. Commands refer to high-level operations to be executed on demand, as detailed in Section\re{sec:usage}.
\itwodlv acts as \textsc{OverGrounder} module and enables incrementality in the computation on the grounding side, while the integrated \textsc{Solver} sub-system currently relies on the same non-incremental solving algorithms adopted in \dlvdue. 
Differently from \dlvdue, in \system, both the grounding and the solving sub-systems are kept alive across the shots; while \itwodlv was already designed to this extent, the \textsc{Solver} module has been modified to remain alive as well. The updates in the solver and the tighter coupling of the two sub-systems pave the way to further steps towards a fully integrated incremental solving.


Multi-shot reasoning is performed by loading a fixed program $P$ at first, and then a set of facts $F_{i}$ for each shot $i$.
According to the techniques described in Section~\ref{sec:overgrounding}, the \textsc{OverGrounder} module maintains across all shots a monotonically growing propositional program $G$. 
Such program is updated at each shot $i$ with the new ground rules generated on the basis of facts in $F_i$ that were never seen in previous shots; then, an internal component efficiently manages ad-hoc internal data structures, updated from shot to shot, that allow to keep track and select a relevant, yet smaller, portion of $G$ to be passed to the \textsc{Solver} module. 
On the basis of such \quo{relevant} portion, the \textsc{Solver} module is then able to 
compute the answer sets of $AS(P \cup F_i)$ for the shot $i$. Note that \quo{relevant} is here used in the sense of \quo{sufficient to keep equivalence with $P \cup F_i$}.
Eventually, once the answer sets are provided as output, the internal data structures of the \textsc{Solver} module are cleaned up from shot-dependent information, so to be ready for the subsequent evaluations. 


%
%
\algdef{SE}[DOWHILE]{Do}{doWhile}{\algorithmicdo}[1]{\algorithmicwhile\ #1}%
\renewcommand{\algorithmicrequire}{\textbf{Input:}}
\renewcommand{\algorithmicensure}{\textbf{Output:}}
\algnewcommand\algorithmicupdates{\textbf{Updates:}}
\algnewcommand\Update{\item[\algorithmicupdates]}
\algrenewcommand{\algorithmiccomment}[1]{\hfill$/\!\!/${\em #1}}


\begin{figure}[t!]
\begin{small}
\begin{algorithmic}[1]
\Require{Non-ground program $P$, ground program $G$, input facts $F_i$ for shot $i$}
\Ensure{A desimplified and enlarged ground program G' = $\DG \cup \NR$}
\Update{the set of deleted rules $D$, collection of set $\UF$ and $\SF$}
\Function{\incrinst}{$P,G,F$}
\State $\DG=G$,
\State $\NR = \emptyset$, $\NF = F_i \setminus \UF$, $\OF = \SF \setminus F_i$
\State $\UF = \UF \cup F_i$, $\SF = \SF \cap F_i$
\While{$\NR \cup \NF$ or $\OF$ have new additions}

 \State{// \em \desimpl\ step}
    
    \State{Undoes simplifications in $DG$;} 
    
    \State{Might move rules from $D$ to $\NR$, and} 
    
    \State{add previously deleted atoms from rules in $DG$;}

\State{// \em \deltainst step}
\Do
   %
   %
   \ForAll{$r \in P$}
        \State{ $I_r = getInstances(r,DG,NR)$ }
        \State{ $I'_r = simplify(I_r)$ }
        \State{ $\NR = \NR \cup I'_r$ }
   \EndFor
   \doWhile{there are additions to $\NR$}
\EndWhile
\State{\Return $G' = \DG \cup \NR$}
\EndFunction
\end{algorithmic}
\caption{\label{alg:simplinf} Simplified version of the incremental algorithm \incrinst}
\end{small}

\end{figure}


\subsection{Implementation Details}\label{subsec:implementationDetails}

%

The evaluation order taking place in the \textsc{OverGrounder} module is carried out by considering direct and indirect dependencies among predicates in $P$. Connected components in the obtained dependency graph are identified once and for all before at the beginning of the first shot: then, the incremental grounding process takes place on a per component basis, following a chosen order. 
We report an abstract version of our \incrinst algorithm in Figure~\ref{alg:simplinf}, where, for the sake of simplicity, we assume the input program $P$ forms a single component.

At shot $i$, a new overgrounded program $G' = DG \cup \NR$ is obtained from $G$ by iteratively repeating, until fixed point, a {\desimpl step}, followed by an {\em Instantiate and Simplify} step, which we call \deltainst. A set $AF$ of {\em accumulated atoms} keeps tracks of possibly true ground atoms found across shots; the set $\NR$ keeps track of newly added rules whose heads can be used to build additional ground rules at current shot, while $DG$ is a \quo{desimplified} version of $G$. 

The {\desimpl} step properly undoes all simplifications applied on $G$ at previous shots that are no longer valid according to $F_i$.
This step relies on the meta-data collected during the previous simplifications: intuitively, meta-data keep record of the ``reasons'' that led to  simplifications, such as deleted rules and/or literals. As a result of this phase, deleted rules might be reinstated, simplified rules might be lengthened, and new additions to $\NR$ could be triggered.

The following {\deltainst} step incrementally processes each rule $r \in P$, possibly producing new additions to $\NR$.
The $getInstances$ function generates all the new ground instances $I_r$ of a rule $r$ by finding substitutions for the variables of $r$ obtained by properly combining head atoms of $DG$ and of $\NR$. 

The instantiation of a rule relies on a version of the classic semi-na\"ive strategy of~\cite{DBLP:books/cs/Ullman88}. The reader may see the work \cite{DBLP:conf/birthday/FaberLP12} for details about a specialized implementation for ASP. The rules of $P$ are processed according to an order induced by predicate dependencies.

Then we simplify $I_r$ to $I'_r$.
In particular, this step processes rules in $I_r$ to check if some can be simplified or even eliminated (see Section~\ref{sec:overgrounding}), still guaranteeing semantics.
On the grounding side, all ground rules in $I_r$ are stored in their complete and non-simplified version along with information (i.e., meta-data) regarding body literals that were simplified, and regarding deleted rules, e.g., those rules containing a certainly false literal in their body.
Then $I'_r$ is added to $\NR$.

It must be noted that $DG$ is subject only to additions and desimplifications, while simplifications are allowed only on the newly added rules $\NR$. Further details on the tailored overgrounding process can be found in~\cite{DBLP:journals/tplp/IanniPZ20}.



Once these two phases are over, the obtained ground rules are used to update the overgrounded program $G$; meta-data related to the occurred simplifications are maintained in order to undo no longer valid simplifications in later shots.
 

Note that $G$ is cumulatively computed across the shots and kept in memory, ready to be re-adapted and possibly enlarged, yet becoming
more generally applicable to a wider class of sets of input facts;
this comes at the price of a generally larger memory footprint. 
Moreover, when fed to the {\sc Solver} module, the size of $G$ can highly influence performance. 
In order to mitigate the latter issue, \system makes use of the aforementioned meta-data for identifying a smaller yet equivalent ground program, which is in turn given as input to the {\sc Solver} module in place of the whole $G$.

Furthermore, to mitigate memory consumption, the system has been endowed with a simple {\em forgetting} strategy that, upon request, removes all rules accumulated in $G$ so far, while still keeping atoms stored; this will cause the overgrounded program to be computed from scratch from the next shot on, but allows one to instantly reduce the memory footprint of the system.
Other finer-grained forgetting strategies for overgrounded programs are presented in~\cite{DBLP:conf/padl/CalimeriIPPZ24}.


\section{System Usage}\label{sec:usage}

\system can be executed either remotely or locally. 
In case of a remote execution, clients can request for a connection specified via an IP address and a port number, corresponding to the connection coordinates at which the system is reachable.
Once a connection is established, the system creates a working session and waits for incoming XML statements specifying which tasks have to be accomplished. The system manages the given commands in the order they are provided.
The possible commands are: {\em Load}, {\em Run}, {\em Forget}, {\em Reset}, and {\em Exit}.

The system works on the assumption that a fixed program $P$ can be loaded once at the beginning of the system's life-cycle; 
multiple set of facts, each representing a specific shot, can be repeatedly loaded; a shot $k$ composed of facts $F_k$ can be {\em run}, i.e. one can ask the system to compute $AS(P \cup F_k)$. Of course, $P$ can possibly contain also facts, which will assumed to be fixed across shots, in contrast with the set $F_k$ which can vary from shot to shot.

The available commands are detailed next:

\paragraph{Load.} 
A \texttt{load} tag can be formed in order to requests to load a program or a set of facts from a file.
The attribute \texttt{path} can be used to specify a string, representing a file path containing what has to be loaded. Multiple load commands can be provided: rules files are accumulated to form a unique program and, similarly, also facts are accumulated.
For instance, with the following commands the system is asked to load four files: 
\begin{center}
\begin{tabular}{l}
\texttt{<load path="my\_rule1.asp"/>}\\
\texttt{<load path="my\_rule2.asp"/>}\\
\texttt{<load path="my\_facts1.asp"/>}\\
\texttt{<load path="my\_facts2.asp"/>}\\
\end{tabular}
\end{center}
Assuming the first two loaded files contain rules while the latter two contain facts, the system composes a fixed program $P$ consisting of all rule files and stores all loaded facts which together compose the first shot's input $F_1$. Note that set of rules can be loaded only at the beginning of the system's activity, while input facts can be loaded at any time.

\paragraph{Run.}
The \texttt{<run/>} command requests to compute the answer sets of the loaded program together with the collected facts. 
As a side effect, incremental grounding takes place, thus updating the current overgrounded program $G_P$.


After a \texttt{run} command is executed, all so far loaded facts are assumed to be no longer true.
Future loading of rule files after the first \texttt{run} are discarded, as $P$ is assumed to be fixed; conversely, one expects further loading of facts forming subsequent shots inputs.

\paragraph{Forget.} 
Since $G_P$ tends to be continuously enlarged, forgetting can be used to save memory by dropping off parts of the accumulated ground program. \system features a form of forgetting which is accessible either as a command or using program annotations.

With the command \texttt{<forget type=``mode''/>}, it is possible to request the ``forgetting'' (i.e., removal) of accumulated atoms or rules along the shots.
More in detail, \texttt{mode} can be either \texttt{r} or \texttt{p} to enable the so-called ``rule-based forgetting''  or ``predicate-based forgetting'', respectively (see~\cit{DBLP:conf/padl/CalimeriIPPZ24}).
The rule-based forgetting removes all ground rules composing the so far accumulated overgrounded program $G_P$, whereas the predicate-based forgetting removes all ground rules and accumulated atoms of all predicates appearing either in rule bodies or heads.  
Overgrounding is started from scratch at the next shot.  
Note that the \texttt{forget} command allows to choose in which shot forgetting happens, but one cannot select which parts of $G_P$ must be removed.

Alternatively, forgetting can be managed by adding \textit{annotations} within $P$. Annotations consist of meta-data embedded in comments (see\cit{DBLP:journals/ia/CalimeriFPZ17}), and allow to specify which predicates or rules have to be forgotten at each shot.
Syntactically, all annotations start with the prefix ``$\%@$'' and end with a dot (``$.$'').
The idea is borrowed from Java and Python annotations having no direct effect on the code they annotate yet allowing the programmer to inspect the annotated code at runtime, thus changing the code behavior at will.
In order to apply the predicate-based forgetting type after each shot over some specific predicates, the user can include in the loaded logic program an annotation of the following form:
\begin{center}
    \texttt{\%@global\_forget\_predicate(p/n).} 
\end{center}
forcing the system to forget all the atoms featuring the predicate \texttt{p} of arity \texttt{n}. 
To forget more than one predicate, the user can simply specify more than one annotation of this type.
Furthermore, an annotation in the form:
\begin{center}
    \texttt{\%@rule\_forget().}
\end{center}
can be used in the logic program before a rule to express that all ground instances of such rule have to be dropped at each shot. The user can annotate more than one rule; each one needs to be preceded by the annotation.

\paragraph{Service commands.}
Further appropriate service commands allow managing the behavior of the system. 
The \texttt{<reset/>} command requests to hard reset all internal data structures, including $P$ and $G_P$, and restarts the computation from scratch, while 
the \texttt{<exit/>} command requests to close the working session and to stop the system.

\medskip

The default reasoning task of \system is the search for just one answer set; it is possible to compute all the existing answer sets with a dedicated switch (option \texttt{$-$n0}).
Alternatively, the system can perform grounding only, and output just the current overgrounded program, which can be piped to a solver module of choice (option \texttt{$--$mode=idlv $-$t}). 

\medskip
\begin{figure}[t]
  \centering
  \resizebox{\textwidth}{!}{
  \begin{tabular}{   m{4.0cm} | m{4.0cm} | m{4.0cm}  }
    \hline
    \begin{minipage}{.30\textwidth}
      \includegraphics[width=\linewidth]{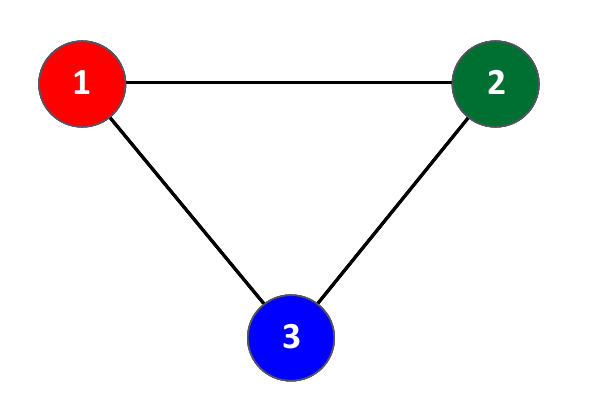}
    \end{minipage}
    &
    \begin{minipage}{.30\textwidth}
      \includegraphics[width=\linewidth]{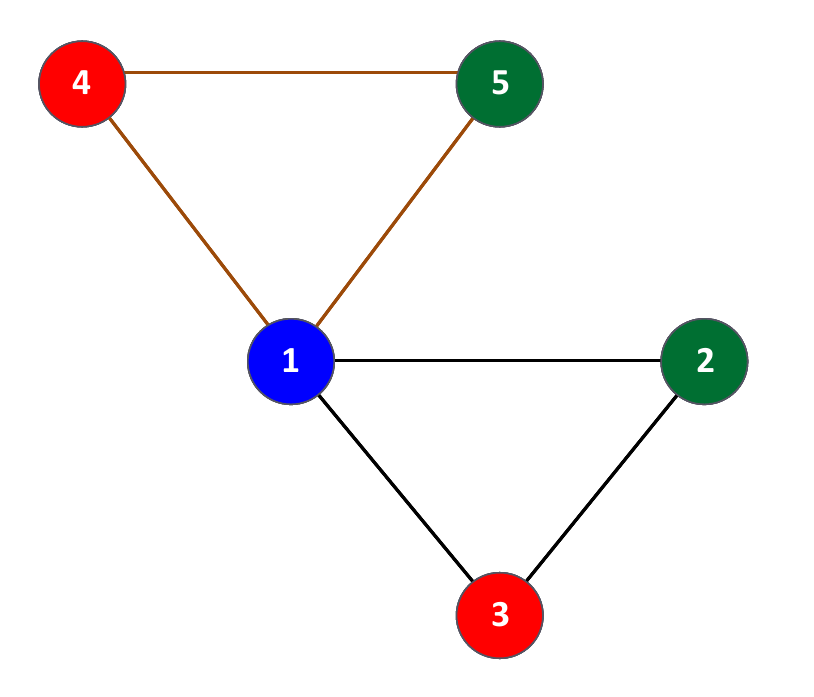}
    \end{minipage}
    &
    \begin{minipage}{.30\textwidth}
      \includegraphics[width=\linewidth]{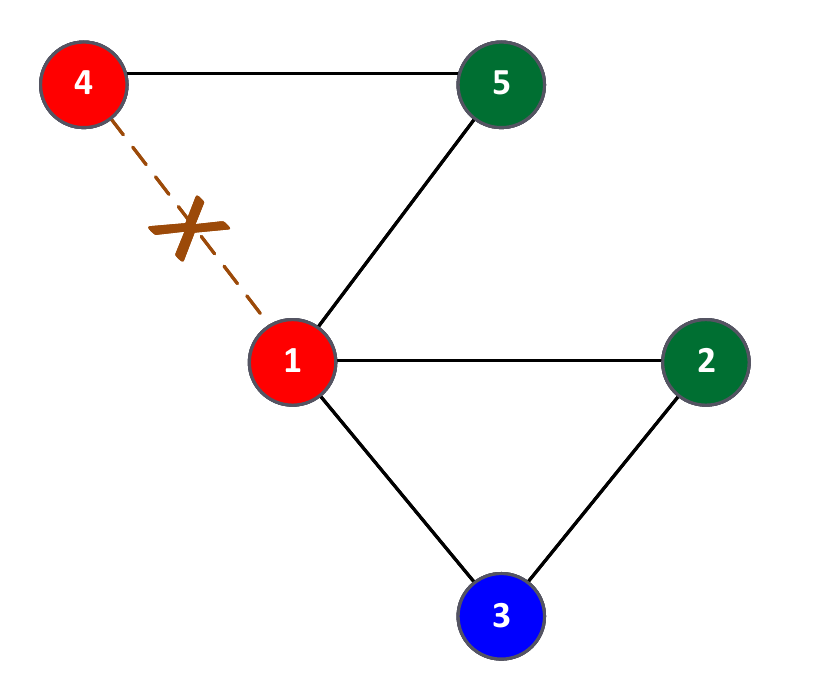}
    \end{minipage}
    \\
    \begin{minipage}{.30\textwidth}
      {Shot 1 - Optimal answer set based on initial configuration (input at shot 1): $F_1$ = \{node(1..3), edge(1,2), edge(2,3), edge(1,3)\}. \\ \phantom{x} \\ } 
    \end{minipage}  
    &
    \begin{minipage}{.30\textwidth}
      {Shot 2 - Optimal answer set based on updated facts: \\
      $F_2$ = $F_1$ $\cup$ \{node(4..5), col(4,red), edge(4,5), \\ edge(1,5), edge(1,4)\}. \\ 
      \\ } 
    \end{minipage} 
    &
    \begin{minipage}{.30\textwidth}
      {Shot 3 - Optimal answer set based on updated facts: \\ $F_3$ = $F_2$ $\setminus$ \{edge(1,4)\}. \\ \phantom{x} \\ }
    \end{minipage}
    \\
    \hline
  \end{tabular}
  }
  \vspace{0.5ex}
  \caption{An example of multi-shot reasoning task based on \system: compute 3-colouring for graphs featuring a structure that changes over time.}\label{fig:3col-example}
\end{figure}
We illustrate next how the system works when properly executed for a multi-shot reasoning task; to this aim, we make use of an example over dynamic graphs, i.e. graphs whose shape changes over time. Dynamic graphs have practical relevance in many real-world scenarios, e.g., communication networks, VLSI design, graphics, assembly planning, IoT, etc. (see\cit{DBLP:reference/crc/DemetrescuFI04a,DBLP:journals/dase/WangYMW19,DBLP:conf/isaac/HeTZ14,zigbee_9431875}); for the sake of presentation, we will consider here a simple setting based on the classical NP-hard 3-colouring problem (\cite{Lawler1976ANO}) in a dynamic setting.  
Given a graph $G(V,E)$, the problem is to assign each node $v\in V$ with exactly one colour out of a set of $3$ (say red, green and blue), so that any pair of adjacent nodes never gets the same colour. 
If the structure of a given graph instance $G$ is specified by means of facts over predicates $node$ and $edge$, then the following program $P_{3col}$ encodes the problem in ASP: 

\begin{center}
\begin{tabular}{l}
$r_1 :\ \ \ col(X, red) \veel col(X, green) \veel col(X, blue)\ \derives\ node(X). $ \\
$r_2 :\ \ \ \derives\ edge(X,Y),\ col(X, C),\ col(Y, C).$ \\
\end{tabular}
\end{center}

Here, $r_1$ is a ``guessing'' rule, expressing that each node must be assigned with one of the three available colours, whereas $r_2$ is a strong constraint that filters out all candidate solutions that assign two adjacent nodes with the same colour. 
We refer here to an optimization version of the problem, in which some preferences over admissible solutions are given; this can be easily expressed via the following {\em weak constraints} (see~\cite{DBLP:journals/tplp/CalimeriFGIKKLM20} for details on the linguistic features of ASP):
\begin{center}
\begin{tabular}{l}
$r_3 :\ \ \ :\sim\ \naf\ col(1,red).\ [1@1]$ \\
$r_4 :\ \ \ :\sim\ \naf\ col(2,green).\ [1@1]$
\end{tabular}
\end{center}
In this case, rules $r_3$ and $r_4$ express preferences for colours to be assigned to nodes $1$ and $2$: more in detail, colour red is preferred for node $1$, while colour green is preferred for node $2$.

Let us consider the setting in which it is needed to reason on a graph whose structure changes over time, that is, at each shot, nodes and edges can be added or removed.

We show the behaviour of the system across three possible shots where input facts change (see Figure\re{fig:3col-example}).
Assuming that rules $r_1$, $r_2$, $r_3$, $r_4$ are contained in a file \texttt{3-col.asp}, the command \texttt{<load path="3-col.asp"/>} has issued to \system to load the program.
Let \texttt{f1.asp} be a file containing the input facts for the first shot: 
\begin{center}
\noindent $node(1..3).\ edge(1,2).\ edge(2,3).\ edge(1,3).$ \\
\end{center}
By issuing the commands \texttt{<load path="f1.asp"/>} and \texttt{<run/>}, the system finds the unique optimum answer set that assigns nodes $1,2,3$ with colours red, green and blue, respectively, and internally stores the overgrounded program $G_{3col}$ reported below, in which barred atoms represent occurred simplifications:
\begin{center}
\begin{tabular}{l}
$r_{1.}\ \ col(1, red) \veel col(1, green) \veel col(1, blue) \derives \hcancel{node(1)}.$ \\
$r_{2.}\ \ col(2, red) \veel col(2, green) \veel col(2, blue) \derives \hcancel{node(2)}. $ \\
$r_{3.}\ \ col(3, red) \veel col(3, green) \veel col(3, blue) \derives \hcancel{node(3)}. $ \\
$r_{4.}\ \ \derives\ \hcancel{edge(1,2),}\ col(1, red),\ col(2, red).$ \\
$r_{5.}\ \ \derives\ \hcancel{edge(1,2),}\ col(1, green),\ col(2, green).$ \\
$r_{6.}\ \ \derives\ \hcancel{edge(1,2),}\ col(1, blue),\ col(2, blue).$ \\
$r_{7.}\ \ \derives\ \hcancel{edge(2,3),}\ col(2, red),\ col(3, red).$ \\
$r_{8.}\ \ \derives\ \hcancel{edge(2,3),}\ col(2, green),\ col(3, green).$ \\
$r_{9.}\ \ \derives\ \hcancel{edge(2,3),}\ col(2, blue),\ col(3, blue).$ \\

$r_{10.}\  \derives\ \hcancel{edge(1,3),}\ col(1, red),\ col(3, red).$ \\
$r_{11.}\  \derives\ \hcancel{edge(1,3),}\ col(1, green),\ col(3, green).$ \\
$r_{12.}\ \derives\ \hcancel{edge(1,3),}\ col(1, blue),\ col(3, blue).$ \\

$r_{13.}\ :\sim\ \naf\ col(1,red).\ [1@1]$ \\
$r_{14.}\ :\sim\ \naf\ col(2,green).\ [1@1]$ 
\end{tabular}
\end{center}

%
At this point, facts in \texttt{f1.asp} are no longer assumed to be true, and the system is ready for a further shot. 
Suppose that now a further file \texttt{f2.asp}  containing the facts for the second shot is loaded (Figure~\ref{fig:3col-example}, middle column): 
\begin{center}
\noindent 
$node(1..3).\ edge(1,2).\ $ $edge(1,3).\ $ $edge(2,3).\ $ \\ 
\noindent
$node(4..5).\ $ $col(4,red).\ $ $edge(4,5).\ edge(1,5).\ edge(1,4).$ \\
\end{center}
 two new nodes connected to each other are added and connected also to node $1$, while colouring for node $4$ is already known to be red.
If now another \texttt{<run/>} command is issued, thanks to the overgrounding-based instantiation strategy, the system only generates further ground rules due to newly added nodes and edges, and adds them to $G_{3col}$: 
\begin{center}
\begin{tabular}{l}
$r_{15.}\  col(4,red) \veel col(4,green) \veel col(4,blue) \derives \hcancel{node(4)}.$\\
$r_{16.}\  col(5,red) \veel col(5,green) \veel col(5,blue) \derives \hcancel{node(5)}.$\\

$r_{17.}\  \derives\ \hcancel{edge(1,4),}\ col(1, red),\ \hcancel{col(4, red)}.$ \\
$r_{18.}\  \derives\ \hcancel{edge(1,4),}\ col(1, green),\ col(4, green).$ \\
$r_{19.}\  \derives\ \hcancel{edge(1,4),}\ col(1, blue),\ col(4, blue).$ \\

$r_{20.}\  \derives\ \hcancel{edge(1,5),}\ col(1, red),\ col(5, red).$ \\
$r_{21.}\  \derives\ \hcancel{edge(1,5),}\ col(1, green),\ col(5, green).$ \\
$r_{22.}\  \derives\ \hcancel{edge(1,5),}\ col(1, blue),\ col(5, blue).$ \\

$r_{23.}\  \derives\ \hcancel{edge(4,5),}\ \hcancel{col(4, red),}\ col(5, red).$ \\
$r_{24.}\  \derives\ \hcancel{edge(4,5),}\ col(4, green),\ col(5, green).$ \\
$r_{25.}\  \derives\ \hcancel{edge(4,5),}\ col(4, blue),\ col(5, blue).$ \\
\end{tabular}
\end{center}

Notably, simplifications made at shot 1 remain valid since all facts in shot $1$ are also facts of shot $2$.
Finally, suppose that in the third shot the system loads a file \texttt{f3.asp} containing the facts (Figure~\ref{fig:3col-example}, right column): 
\begin{center}
\noindent 
$node(1..3).\ edge(1,2).\ $ $edge(2,3).\ $ $edge(1,3).\ $ \\ 
\noindent
$node(4..5).\ $ $col(4,red).\ $ $edge(1,5).\ edge(4,5).$ \\
\end{center}
The input graph is updated by removing the edge between nodes $1$ and $4$. 
Now, no new fact results as unseen in previous shots: hence, no additional ground rules are generated and no grounding effort is needed; the only update in $G_{3col}$ consists of the desimplification of $edge(1,4)$ in rules $r_{17}$, $r_{18}$ and $r_{19}$:

\begin{center}
\begin{tabular}{l}
$r_{15.}\  col(4,red) \veel col(4,green) \veel col(4,blue) \derives \hcancel{node(4)}.$\\
$r_{16.}\  col(5,red) \veel col(5,green) \veel col(5,blue) \derives \hcancel{node(5)}.$\\

$r_{17.}\  \derives\ edge(1,4),\ col(1, red),\ \hcancel{col(4, red)}.$ \\
$r_{18.}\  \derives\ edge(1,4),\ col(1, green),\ col(4, green).$ \\
$r_{19.}\  \derives\ edge(1,4),\ col(1, blue),\ col(4, blue).$ \\

$r_{20.}\  \derives\ \hcancel{edge(1,5),}\ col(1, red),\ col(5, red).$ \\
$r_{21.}\  \derives\ \hcancel{edge(1,5),}\ col(1, green),\ col(5, green).$ \\
$r_{22.}\  \derives\ \hcancel{edge(1,5),}\ col(1, blue),\ col(5, blue).$ \\

$r_{23.}\  \derives\ \hcancel{edge(4,5),}\ \hcancel{col(4, red),}\ col(5, red).$ \\
$r_{24.}\  \derives\ \hcancel{edge(4,5),}\ col(4, green),\ col(5, green).$ \\
$r_{25.}\  \derives\ \hcancel{edge(4,5),}\ col(4, blue),\ col(5, blue).$ \\
\end{tabular}
\end{center}

The unique optimum answer set now consists in colouring nodes $1,2,3,4,5$ in red, green, blue, red, green, respectively.
It is worth noting that the savings in computational time, obtained by properly reusing ground rules generated at previous shots, occur with no particular assumption made in advance about possible incoming input facts. Moreover, the management of the incremental computation is completely automated and transparent to the user, who is not required to define a priori what is fixed and what might change.

\section{Experimental Analysis}\label{sec:exp+disc}
In this section we discuss the performance of \system when executing multi-shot reasoning tasks in real-world scenarios.

\subsection{Benchmarks}\label{sec:bench}
We considered a collection of real-world problems that have been already used for testing incremental reasoners. 
A brief description of each benchmark follows.
The full logic programs, instances and experimental settings can be found at \textit{\url{https://dlv.demacs.unical.it/incremental}}.
\medskip

\paragraph{\pacman (\cite{DBLP:conf/ruleml/CalimeriGIPPZ18}).}
This domain models the well-known real-time game \pacman. 
Here, a logic program $P_{pac}$ describes the decision-making process of an artificial player guiding the \pacman in a real implementation. 
The logic program $P_{pac}$ is repeatedly executed together with different inputs describing the current status of the game board. The game map is of size $30 \times 30$, and includes the current position of enemy ghosts, the position of pellets, of walls, and any other relevant game information.
Several parts of $P_{pac}$ are \quo{grounding-intensive}, like the ones describing the distances between different positions in the game map. These make use of a predicate $distance(X_1,Y_1, X_2, Y_2,D)$, where $D$ represents the distance between points $(X_1,Y_1)$ and $(X_2,Y_2)$, obtained by taking into account the shape of the labyrinth in the game map.

\medskip

\paragraph{\mmedia (\cite{DBLP:journals/tplp/IanniPZ20}).}
This domain is obtained from the multi-media video streaming context (see~\cite{DBLP:conf/icc/BeckBDEHS17}).
In this scenario, one of the common problems is to decide the caching policy of a given video content, depending on variables like the  number and the current geographic distribution of viewers. 
The caching policy is managed via a logic program $P_{cc}$. 
In particular, policy rules are encoded in the answer sets $AS(P_{cc} \cup E)$, where $E$ encodes a continuous stream of events describing the evolving popularity level of the content at hand. 
This application has been originally designed in the LARS framework of~\cite{DBLP:journals/ai/BeckDE18}, using time window operators in order to quantify over past events. 
We adapted the available LARS specification according to the conversion method presented by~\cite{DBLP:journals/tplp/BeckEB17} to obtain $P_{cc}$ as a plain logic program under answer set semantics; events over $30\,000$ time points were converted to corresponding sets of input facts.
This category of stream reasoning applications can be quite challenging, depending on the pace of events and the size of time windows. 

\medskip

\paragraph{\pvsystem (\cite{idlvsr}).} 
We consider here a stream-reasoning scenario in which an Intelligent Monitoring System (IMS) for a photo-voltaic system (PVS) is used to promptly detect major grid malfunctions. 
We consider a PVS composed by a grid of $60 \times 60$ solar panels interconnected via cables; each panel continuously produces a certain amount of energy to be transferred to a Central Energy Accumulator (CEA), 
directly or via a path between neighbour panels across the grid. 
The amount of energy produced is tracked and sent to the IMS. 
An ASP program is repeatedly executed over streamed data readings with the aim of identifying situations to be alerted for, and thus prompting the necessity of maintenance interventions. 
Notably, this domain causes a more intensive computational effort on the grounding side with respect to the solving side as the logic program at hand does not feature disjunction and is stratified.

\begin{figure*}[t]
\begin{subfigure}{.495\textwidth}
  \includegraphics[width=\linewidth]{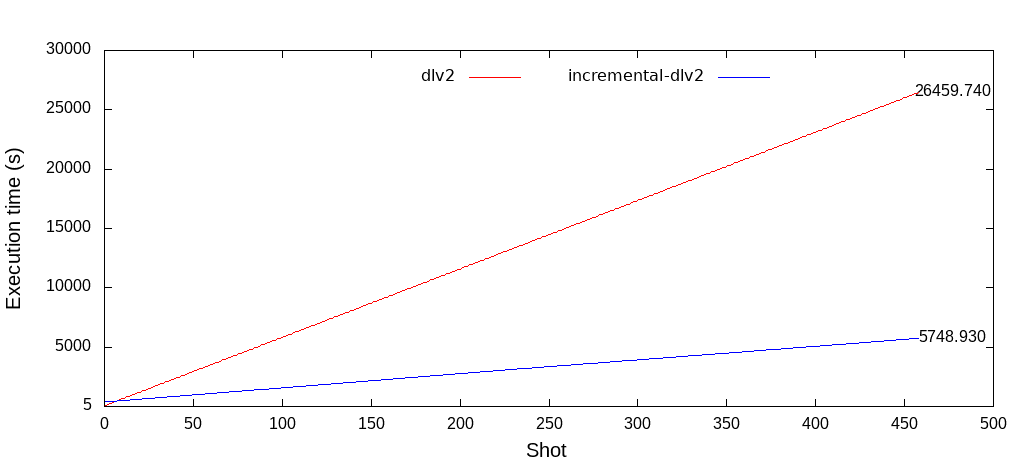}
  \caption{\pacman: cumulated time}\label{fig:pacman-times}
\end{subfigure}
\begin{subfigure}{.495\textwidth}
  \includegraphics[width=\linewidth]{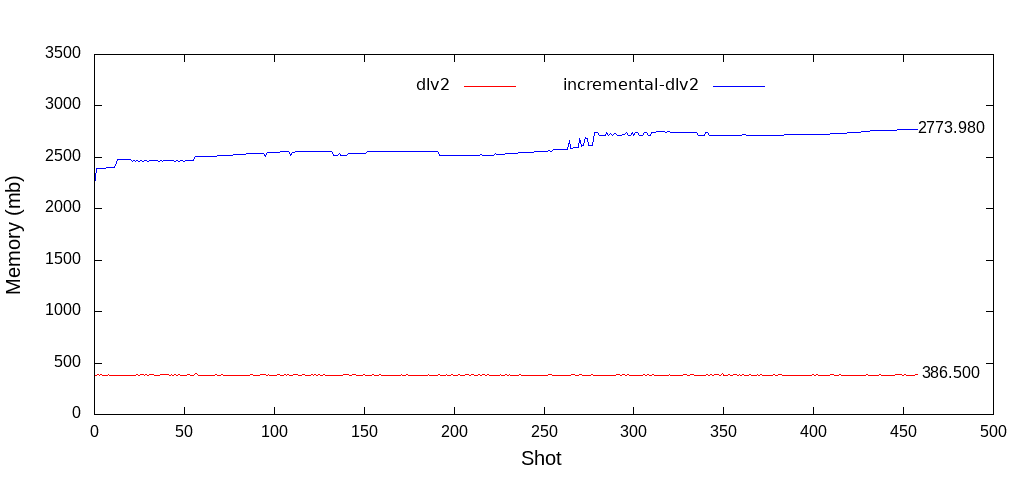}
  \caption{\pacman: memory usage}\label{fig:pacman-memory}
\end{subfigure}
\begin{subfigure}{.495\textwidth}
  \includegraphics[width=\linewidth]{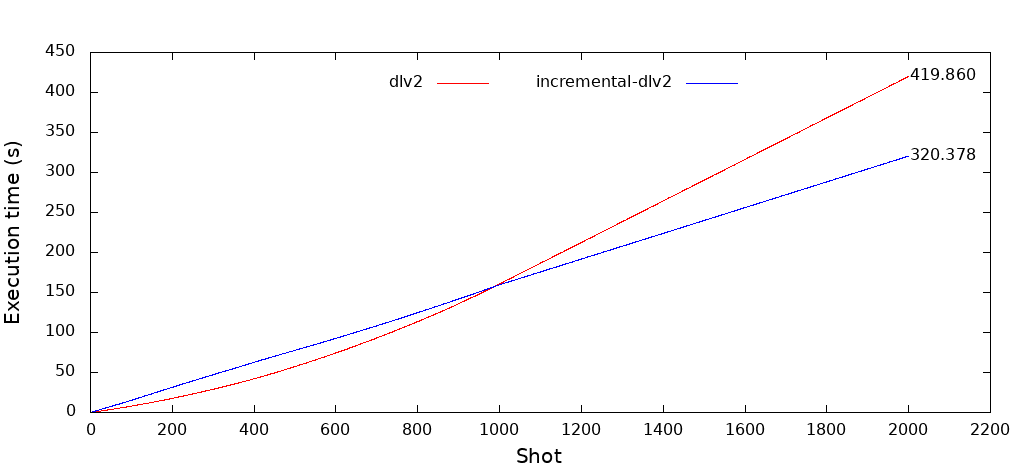}
  \caption{\mmedia: cumulated time}\label{fig:mmedia-times}
\end{subfigure}
\begin{subfigure}{.495\textwidth}
  \includegraphics[width=\linewidth]{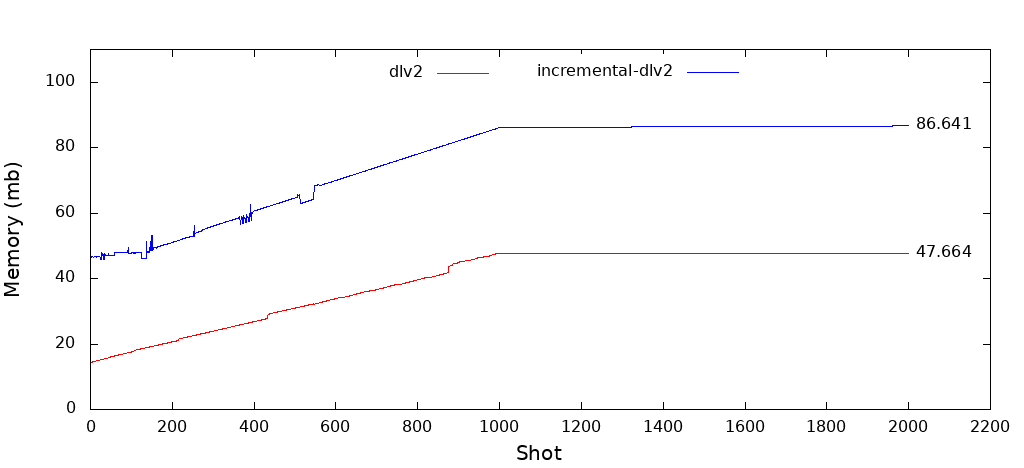}
  \caption{\mmedia: memory usage}\label{fig:mmedia-memory}
\end{subfigure}
\begin{subfigure}{.495\textwidth}
  \includegraphics[width=\linewidth]{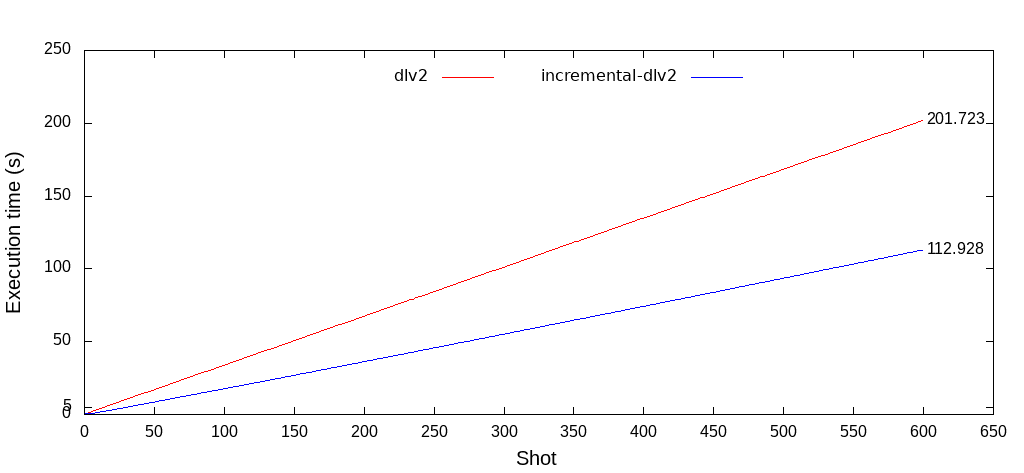}
  \caption{Photo-Voltaic System: cumulated time}\label{fig:pvsystem-times}
\end{subfigure}
\begin{subfigure}{.495\textwidth}
  \includegraphics[width=\linewidth]{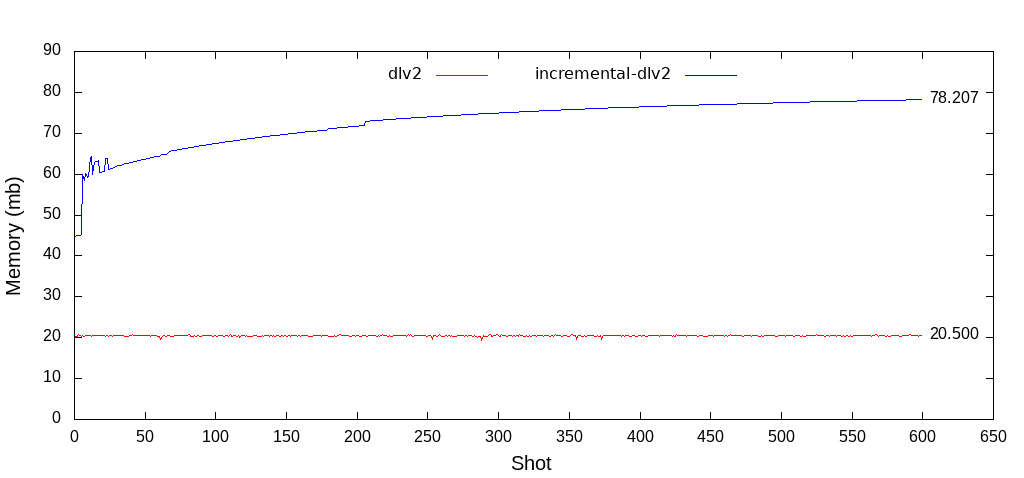}
  \caption{Photo-Voltaic System: memory usage}\label{fig:pvsystem-memory}
\end{subfigure}
\caption{\system against \dlvdue: performance in multi-shot contexts.}\label{fig:exp-results}
\end{figure*}

\subsection{Setting}

We compared the herein presented \system system against the \dlvdue system.
Both systems were run in single-threaded mode. Experiments have been performed on a NUMA machine equipped with two $2.8$GHz AMD Opteron $6320$ CPUs, with $16$ cores and $128$GB of RAM. 
Differently from \system, \dlvdue is restarted, i.e., executed from scratch, at each shot in order to evaluate the given program on the current facts.  
For each domain, we choose two different measures: we track the total accumulated time and the maximum memory peak per shot.
Figure\re{fig:exp-results} plots the chosen measures against the number of shots for all benchmarks; the $X$ axis diagrams data in the shot execution order.


\subsection{Execution times}

When observing execution times, a clear advantage is experienced by \system over all considered domains.

The \pacman benchmark is characterized by the need for computing all distances between all positions ($30 \times 30$) in the game map, which is fixed, and hence \system is able to compute them only once at the first shot, which is the most time expensive. 
At each shot after the first, there is a net gain in grounding times stabilizing at an approximate speedup factor of 4.6; furthermore, given the nature of the available instances, which encode a real game, it turns out that, for a large part of the overgrounded program, the simplifications performed are preserved when moving towards next shots.
Hence, \system gains from the low need for desimplifications, while still avoiding to burden the {\sc Solver} module too much as it is fed with highly simplified versions of the ground program.

In the \mmedia scenario, reasoning is performed on a time window spanning over the last 1000 time ticks. Thus, instances are so that for the first $1000$ shots the input facts span over a time window of less than $1000$ ticks and consequently, they are a superset of inputs coming from previous shots; then, from shot $1000$ on, possible input facts do not change anymore. 
This implies that, basically, from shot to shot the whole overgrounded program is just monotonically enriched with new ground rules, yet keeping the simplifications performed in previous shots, up to the point that nothing new has to be instantiated after shot $1000$.
While \system cumulative time performance exhibits a linear growth, the \dlvdue behaviour worsens as instances become larger. 
Indeed, early shots require a little computational effort and thus \dlvdue takes advantage from the lack of the multi-shot machinery overhead, although the computing times of the two systems are very close.
However, in later shots the picture overturns: the growth of \dlvdue times starts accelerating after a few hundred shots, up to the point that, from shot $1000$ on, as the task to be executed is almost the same across all shots, \system significantly outperforms \dlvdue, as it basically saves the whole grounding time thanks to the overgrounding technique. 
It is also worth noting that \system scales definitely better, as the growth of the execution time is always almost linear; on the other hand, the \dlvdue cumulative execution time has a quadratic like trend at the very start, and becomes linear only when the effort for requested tasks stabilizes after shot $1000$, converging to a speedup factor in favor of \system of around 1.6.

In the case of the \pvsystem benchmark, both systems show a linear growth in cumulated time; still, \system clearly outperforms \dlvdue, with a speedup factor of slightly less than 2.
The corresponding logic program is stratified and features a recursive component that is ``activated'' at each shot; hence, the hard part of the computation is carried out during the grounding phase, which, also given the nature of the available instances, still remains significant in later shots, differently from the other domains.

\subsection{Memory usage}

Some additional considerations deserve to be done about memory usage. 
Indeed, as it is expected because of the incremental grounding strategy herein adopted, 
the memory footprint is definitely higher for \system, in all considered domains.
However, interestingly, it can be noted that in all cases the memory usage trend shows an asymptotic ``saturation'' behaviour: after a certain number of shots the memory usage basically stays constant; hence, the price to pay in terms of memory footprint is not only counterbalanced by the gain in terms of performance, but it also happens to not ``explode''.
We also observe that in the \mmedia and \pvsystem benchmarks the memory usage increases along the shots, while it is reached a sort of plateau in the \pacman benchmark.
Indeed, in this latter domain a large amount of information, useful in all shots, is inferred only at the first shot and then kept in memory, but with some redundancy. 
As a result, the memory usage in this benchmark domain is high at the beginning of the shot series but it stays almost unchanged later. On the other hand \dlvdue makes a good job in generating, from scratch, a compact ground program per each shot.
Although the results show a fairly reasonable memory usage, 
as mentioned before, memory-limiting and rule-forgetting policies added on top of existing algorithms can help 
in mitigating the memory footprint of \system, especially in scenarios where memory caps are imposed.

\section{Related Work}\label{sec:relwork}


\subsection{Theoretical foundations of incremental grounding in ASP}

The theoretical foundations and algorithms at the basis of \system were laid out by~\cite{DBLP:journals/tplp/CalimeriIPPZ19} and~\cite{DBLP:journals/tplp/IanniPZ20}. The two contributions propose respectively a notion of {\em embedding} and {\em tailored embedding}. {\em Embeddings} are families of ground programs which enjoy a number of desired properties. Given $P \cup F$, an embedding $E$ is such that $AS(P \cup F) = AS(E)$; $E$ must be such that it {\em embeds} ($E \embeds r$) all $r \in ground(P \cup F)$. 

The $\embeds$ operator is similar to the operator $\models$ which is applied to interpretations, and enjoys similar model theoretical properties. 
Intuitively, given interpretation $I$ and rule $r$ with head $h_r$ and positive body $b_r$, it is known that $I \models r$ whenever $I \models h_r$ or $I \not\models b_r$, thus enforcing an implicative dependency between $b_r$ and $h_r$. A similar implicative dependency is enforced on the structure of ground programs qualified as embeddings: $E \embeds r$ whenever $r \in E$ or whenever, for some atom $b \in b_r$, $E$ does not embed any rule having $b$ in its head.
Embeddings are closed under intersection and the unique minimal embedding can be computed in a bottom up fashion by an iterated fixed point algorithm. An overgrounded program $G$ of $P \cup F$ is such that $G \cup F$ is an embedding of $P \cup F$.

\cite{DBLP:journals/tplp/IanniPZ20} extend the notion of embedding to {\em tailored embeddings}. Tailored embeddings are families of ground programs, equivalent to some $P \cup F$, which allow the possibility of including in the ground program itself a simplified version $r'$ of a rule $r \in ground(P)$. A tailored embedding is such that $T \tailors r$ for each $r \in ground(P \cup F)$. The operator $\tailors$ takes into account the possibility of simplifications and deletion of rules. The presence of simplified rules might lead to ground programs which are not comparable under plain set containment: however, tailored embeddings are closed under a generalized notion of containment, and the least tailored embedding can be computed using a bottom up fixed point algorithm. Importantly, an overgrounded program with tailoring $G$ obtained by the \incrinst algorithm at shot $i$ with input facts $F_i$, is such that $G \cup F_i$ is a tailored embedding of $P \cup F_i$, and thus $AS(G \cup F_i) = AS(P \cup F_i)$.

It is worth highlighting that embeddings and tailored embeddings can be seen as families of relativized hyperequivalent logic programs in the sense of~\cite{DBLP:journals/tplp/TruszczynskiW09}.
Indeed, given logic programs $P$ and $Q$, these are said to be hyperequivalent relatively to a finite family of programs $\cal{F}$ iff $AS(P \cup F) = AS(Q \cup F)$ for each $F \in \cal{F}$. 
A member $G$ of a sequence of overgrounded programs is characterized by being equivalent to a program $P$ relative to (part of) a finite set of inputs $F_1, \dots, F_n$, similarly to hyperequivalent programs relative to finite sets of inputs. 
Conditions in which a form of equivalence is preserved under simplifications, possibly changing the simplified program signature w.r.t. the original program, were studied by \cite{woltkr23}.


\subsection{Other ASP systems with incremental features}

The ASP system {\em clingo}~\cite{DBLP:journals/tplp/GebserKKS19} represents the main contribution related to multi-shot reasoning in ASP. 
{\em clingo} allows to procedurally control which and how parts of the logic program have to be incremented, updated and taken into account among consecutive shots. 
This grants designers of logic programs a great flexibility; however, the approach requires specific knowledge about how the system internally holds its computation and on how the domain at hand is structured. It must in fact be noted that the notion of \quo{incrementality} in {\em clingo} is intended in a constructive manner as the management of parts of the logic program that can be built in incremental layers. 
Conversely, in the approach proposed in this paper, the ability of using procedural directives is purposely avoided, in favour of a purely declarative approach. Incrementality is herein intended as an internal process to the ASP system, which works on a fixed input program.

%

The Stream Reasoning system {\em Ticker} of~\cite{DBLP:journals/tplp/BeckEB17} represents an explicit effort towards a more general approach to ASP incremental reasoning. 
Ticker implements the LARS stream reasoning formal framework of~\cite{DBLP:journals/ai/BeckDE18}.
The input language of LARS allows window operators, which enable reasoning on streams of data under ASP semantics.
Ticker implements a fragment of LARS, with no disjunction and no constraints/odd-cycle negation loops, by using back-end incremental truth maintenance techniques.


Among approaches that integrate tightly grounding and solving, it is worth mentioning lazy
grounding (see \cite{DBLP:journals/fuin/PaluDPR09,DBLP:journals/tplp/LefevreBSG17,DBLP:conf/aaai/BomansonJW19}). Note that overgrounding is essentially orthogonal to lazy grounding techniques, since these latter aim at blending grounding tasks within the solving step for reducing memory consumption; rather, our focus is on making grounding times negligible on repeated evaluations by explicitly allowing the usage of more memory, while still keeping a loose coupling between the two evaluation steps.

\subsection{Incrementality in Datalog}

The issue of incremental reasoning on ASP logic programs is clearly related to the problem of maintaining views expressed in Datalog. In this respect, \cite{DBLP:journals/ai/MotikNPH19} proposed the so called {\em delete/rederive} techniques, which aim at updating materialized views. In this approach, no redundancy is allowed, i.e. updated views reflect only currently true logical assertions: this differs from the overgrounding idea, which aims to materialize bigger portions of logic programs which can possibly support true logic assertions. \cite{DBLP:journals/ai/HuMH22} extended further the idea, by proposing a general method in which modular parts of a Datalog view can be attached to ad-hoc incremental maintenance algorithms. For instance, one can plug in the general framework a special incremental algorithm for updating transitive closure patterns, etc.

\section{Future Work and Conclusions}
\label{sec:conclusions}
As future work is concerned, we plan to further extend the incremental evaluation capabilities of \system, by making the solving phase connected in a tighter way with grounding, in the multi-shot setting.
%
Moreover, in order to limit the impact of memory consumption, we intend to study new forgetting strategies to be automatic, carefully timed and more fine-grained than the basic ones currently implemented.
Besides helping at properly managing the memory footprint, such strategies can have a positive impact also on performance; think, for instance, of scenarios where input highly varies across different shots: from a certain point in time on, it is very likely that only a small subset of the whole amount of accumulated rules will actually play a role in computing answer sets. 
As a consequence, accumulating rules and atoms may easily lead to a worsening in both time and memory performance: 
here, proper forgetting techniques can help at selectively dropping the part of the overgrounded program that constitutes a useless burden, thus allowing to enjoy the advantages of overgrounding at a much lower cost. A variant of this approach has been proposed by~\cite{DBLP:conf/padl/CalimeriIPPZ24}.
Investigating the relationship between overgrounded programs and the notion of relatived hyperequivalence of\cit{DBLP:journals/tplp/TruszczynskiW09}, possibly under semantics other than the answer set one, deserves further research.

\bibliographystyle{acmtrans}
\bibliography{references.bib}

\newpage


\end{document}